\documentclass[sigconf]{acmart}

\makeatletter
\def\@acmISBN{}
\def\@acmDOI{}
\def\@formatdoi{}
\makeatother

\makeatletter
\def\maketitle{%
  \if@ACM@anonymous
    \ifnum\num@authorgroups=0\author{}\fi
  \fi
  \begingroup
  \let\@footnotemark\@footnotemark@nolink
  \let\@footnotetext\@footnotetext@nolink
  \renewcommand\thefootnote{\@fnsymbol\c@footnote}%
  \global\@topnum\z@ 
  \global\@botnum\z@ 
  \hsize=\textwidth
  \def\@makefnmark{\hbox{\@textsuperscript{\@thefnmark}}}%
  \@mktitle\if@ACM@sigchiamode\else\@mkauthors\fi\@mkteasers
  \@printtopmatter
  \if@ACM@sigchiamode\@mkauthors\fi
  \setcounter{footnote}{0}%
  \def\@makefnmark{\hbox{\@textsuperscript{\normalfont\@thefnmark}}}%
  \@titlenotes
  \@subtitlenotes
  \@authornotes
  \let\@makefnmark\relax  \let\@thefnmark\relax
  \let\@makefntext\noindent
  \ifx\@empty\thankses\else
    \footnotetextcopyrightpermission{%
      \def\par{\let\par\@par}\parindent\z@\@setthanks}%
  \fi
  \footnotetextcopyrightpermission{%
    \if@ACM@authordraft
        \raisebox{-2ex}[\z@][\z@]{\makebox[0pt][l]{\large\bfseries
            Unpublished
            working draft. Not for distribution}}%
       \color[gray]{0.9}%
    \fi
    \parindent\z@\parskip0.1\baselineskip
    \if@ACM@authorversion\else
      \if@printpermission\@copyrightpermission\par\fi
    \fi
    \if@ACM@manuscript\else
       \if@ACM@journal\else 
         {\itshape \acmConference@shortname, \acmConference@date, \acmConference@venue}\par
       \fi
    \fi
    \if@printcopyright
      \copyright\ \@copyrightyear\ \@copyrightowner\\
    \else
     \@copyrightyear.\
    \fi
    \if@ACM@manuscript
      Manuscript submitted to ACM\\
    \else
      \if@ACM@authorversion
          This is the author's version of the work. It is posted here for
          your personal use. Not for redistribution. The definitive Version
          of Record was published in
          \if@ACM@journal
            \emph{\@journalName}%
          \else
            \emph{Proceedings of \acmConference@name, \acmConference@date}%
          \fi
          \ifx\@acmDOI\@empty
          .
          \else
            , \@formatdoi{\@acmDOI}.
          \fi\\
        \else
          \if@ACM@journal
            \@permissionCodeOne/\@acmYear/\@acmMonth-ART\@acmArticle
            \ifx\@acmPrice\@empty\else\ \$\@acmPrice\fi\\
            \@formatdoi{\@acmDOI}%
          \else 
             \@acmISBN
             \ifx\@acmPrice\@empty.\else\dots\$\@acmPrice\fi\\
             \@formatdoi{\@acmDOI}%
          \fi
        \fi
      \fi}
  \endgroup
  \setcounter{footnote}{0}%
  \@mkabstract
  \if@ACM@printccs
    \ifx\@concepts\@empty\else\bgroup
      {\@specialsection{CCS Concepts}%
         \@concepts\par}\egroup
     \fi
   \fi
   \ifx\@keywords\@empty\else\bgroup
      {\if@ACM@journal
         \@specialsection{Additional Key Words and Phrases}%
       \else
         \@specialsection{Keywords}%
       \fi
         \@keywords}\par\egroup
   \fi
  \andify\authors
  \andify\shortauthors
  \global\let\authors=\authors
  \global\let\shortauthors=\shortauthors
  \if@ACM@printacmref
     \@mkbibcitation
  \fi
  \hypersetup{pdfauthor={\authors},
    pdftitle={\@title},
    pdfsubject={\@concepts},
    pdfkeywords={\@keywords}}%
  \@printendtopmatter
  \@afterindentfalse
  \@afterheading
}
\makeatother

\usepackage{booktabs} 

\setcopyright{rightsretained}

\acmConference[CAIR'17]{1st International Workshop on Conversational Approaches to Information Retrieval}{August 11, 2017}{Tokyo, Japan} 
\acmYear{2017}
\copyrightyear{2017}

\acmPrice{}

\usepackage{setspace}

\DeclareMathOperator*{\argmax}{arg\,max}
\DeclareMathOperator{\softmax}{softmax}

\usepackage[export]{adjustbox}
\usepackage{subcaption}
\usepackage{acronym}
\acrodef{IR}{Information Retrieval}
\acrodef{AMN}{Attentive Memory Network}
\acrodef{RNN}{Recurrent Neural Network}
\acrodef{GRU}{Gated Recurrent Unit}
\acrodef{LSTM}{Long Short-Term Memory}

\title[Attentive Memory Networks]{Attentive Memory Networks: Efficient Machine Reading\\ for Conversational Search}

\author{Tom Kenter}
\affiliation{%
\institution{University of Amsterdam}
\city{Amsterdam}
\country{The Netherlands}
}
\email{tom.kenter@uva.nl}

\author{Maarten de Rijke}
\orcid{0000-0002-1086-0202}
\affiliation{%
\institution{University of Amsterdam}
\city{Amsterdam}
\country{The Netherlands}
}
\email{derijke@uva.nl}

\begin{document}

\begin{abstract}
  Recent advances in conversational systems have changed the search paradigm.
  Traditionally, a user poses a query to a search engine that returns an answer based on its index, possibly leveraging external knowledge bases and conditioning the response on earlier interactions in the search session.
  In a natural conversation, there is an additional source of information to take into account: utterances produced earlier in a conversation can also be referred to and a conversational IR system has to keep track of information conveyed by the user during the conversation, even if it is implicit.

  We argue that the process of building a representation of the conversation can be framed as a machine reading task, where an automated system is presented with a number of statements about which it should answer questions.
  The questions should be answered solely by referring to the statements provided, without consulting external knowledge.
  The time is right for the information  retrieval community to embrace this task, both as a stand-alone task and integrated in a broader conversational search setting. 
  
  In this paper, we focus on machine reading as a stand-alone task and present the \ac{AMN}, an end-to-end trainable machine reading algorithm.
  Its key contribution is in efficiency, achieved by having an hierarchical input encoder, iterating over the input only once.
Speed is an important requirement in the setting of conversational search, as gaps between conversational turns have a detrimental effect on naturalness.  
  On 20 datasets commonly used for evaluating machine reading algorithms we show that the \ac{AMN} achieves performance comparable to the state-of-the-art models, while using considerably fewer computations.
\end{abstract}

%
%



\maketitle


\section{Introduction}

Recent advances in conversational systems \cite{serban2016buildinged,radlinski2017theoretical} have changed the search paradigm.
In a classic setting, a search engine answers a query based on an index, possibly enriching it with information from an external knowledge base \cite{wang2015queryut}.
Additionally, previous interactions in the same session can be leveraged \cite{eickhoff2014lessonsft}.
In addition to these sources, in natural language conversations, information contained in previous utterances can be referred to, even implicitly.
Suppose a conversational system has to answer the query \textit{Where are my keys?} based on a previous statement \textit{I was home before I went to work, which is where I found out I didn't have my keys with me}.
The statement conveys a lot of information, including the likely possibility that the keys are still at the speaker's house.
As is clear from this example, indices or external knowledge bases are of no avail in this setting.
It is crucial for a conversational system to maintain an internal state, representing the dialogue with the user so far.
To address this issue, substantial work has been done in goal-oriented dialogues, tailored to specific settings such as restaurant reservations \cite{bordes2016learning} and the tourist domain \cite{kim2017fourth}.
We argue that a generic conversational agent should be able to maintain a dialogue state without being constrained to a particular task with predetermined slots to be filled.
The time has come for the \ac{IR} community to address the task of machine reading for conversational search~\citep{radlinski2017theoretical}.

As an important step towards generic conversational IR~\citep{kiseleva-evaluating-2017}, we frame the task of conversational search as a general machine reading task \citep{hewlett2016wikireading,hermann2015teaching}, where a number of statements is provided to an automated agent that answers questions about it.
This scenario is different from the traditional question answering setting, in which questions are typically factoid in nature, and answers are based on background knowledge or external sources of knowledge.
In the machine reading task, much as in a natural conversation, a number of statements is provided, and the conversational agent should be able to answer questions based on its understanding of these statements alone.
In \citep{hewlett2016wikireading}, for example, a single Wikipedia page is provided to a machine algorithm which has to answer questions about it.
In \citep{weston2015towards} the machine reads stories abouts persons and objects and has to keep track of their whereabouts. 

Memory networks have proven to be an effective architecture in machine reading tasks \cite{sukhbaatar2015endtoendmn,weston2014memory}.
Their key component is a memory module in which the model stores intermediate representations of input, that can be seen as multiple views on the input so far, from which a final output is computed.
Speed is an important constraint in the context of conversational agents, since long pauses between turns hamper the naturalness of a conversation.
We strive for an efficient architecture, and propose to use a hierarchical input encoder.
Input can be large, hundreds of words, and we hypothesize that first processing the input to get a smaller set of higher-level input representations can benefit a network in two ways: (1)~the higher-level representations provide a distilled representation of the input; (2)~as there are fewer higher-level representations it should be (computationally) easier for the network to focus on the relevant parts of the input.
In short, in this paper we present the \acfi{AMN}, an end-to-end trainable memory network, with hierarchical input encoder.
To test its general applicability we use 20 machine reading datasets specifically designed to highlight different aspects of natural language understanding.
We show that the \ac{AMN} achieves performance comparable to the state-of-the-art models, while using considerably fewer computations.


\section{Related work}

Machine reading is a much-studied domain \cite{cheng2016long,hewlett2016wikireading,hermann2015teaching}.
It is related to question answering, the difference being that in question answering, external domain or world knowledge is typically needed to answer questions \cite{yang2015wikiqa,fader2013paraphrasedrivenlf,miller2016keyvaluemn}, while in machine reading answers should be inferred from a given text.

Hierarchical encoders are employed in a dialogue setting in \cite{serban2016hierarchical} and for query suggestion in \cite{sordoni2015hierarchical}.
In both works, the hierarchical encoder is also trained, for every input sentence, to predict every next input sentence, a setting we did not experiment with.

We build on previous work on memory networks \cite{weston2014memory,sukhbaatar2015endtoendmn,tran2016recurrent}, in particular on dynamic memory networks \cite{xiong2016dynamicmn,kumar2016askma}.
Memory networks are an extension of standard se\-quence-to-sequence architectures; their distinguishing feature is a memory module added between the encoder and decoder.
As they are typically applied in question answering settings, there are two encoders, one for a question and one for a document the question is about.
The decoder does not have access to the input but only to the memory module, which distills relevant information from the input, conditioned on the question. 
The key difference between the Attentive Memory Network we propose and the work in \cite{xiong2016dynamicmn,kumar2016askma}, is in the defining component, the memory module.
In \cite{xiong2016dynamicmn,kumar2016askma}, to obtain every next memory, a \ac{GRU} cell iterates over the input sequence.
This leads to a memory intensive and computationally expensive architecture, since multiple cells are repeatedly being unrolled over the input sequence.
The number of steps an \ac{RNN} is unrolled for, i.e., the number of input representations it reads, together with the hidden state size, is the main determining factor regarding computational complexity.
Therefore, we propose to obtain memories by an \ac{RNN} that, rather than iterating over the entire input, only applies attention over it, which is a much cheaper operation (see \S\ref{section:ourMethod}).

In \cite{sukhbaatar2015endtoendmn} an attention-based memory network is presented, where the input is represented as a sequence of embeddings on which attention is computed (i.e., there is no input reader).
Our Attentive Memory Network differs from this work in that we do use an input reader, a hierarchical \ac{RNN}.
As a consequence, our memory module has far fewer hidden states to attend over.
At the output side, we use \acp{GRU} to decode answers, which is different from the softmax over a dot product between the sum of attention-weighted input and question employed in \cite{sukhbaatar2015endtoendmn}.

To sum up, we propose a memory network that shares its overall architecture with previous models, and that differs in how all key components are constructed, with a view to improve efficiency and, thereby, enable its usage in conversational search scenarios.


\section{Attentive memory networks}\label{section:ourMethod}

\acresetall

To facilitate the presentation of our Attentive Memory Networks, we first briefly recapitulate standard sequence-to-sequence models.

\subsection*{Recurrent cells}

An input sequence is processed one unit per time step, where the recurrent cell computes a new state $h_{t}$ as a function of an input representation $x$ and a hidden state $h_{t-1}$ as:
\begin{equation}\label{eq:recurrence}
  \mathbf{h}_{t} = f(\mathbf{x}, \mathbf{h}_{t-1} ; \theta),
\end{equation}
based on internal parameters $\theta$.
The function $f$ itself can be implemented in many ways, for example as an \ac{LSTM} \cite{hochreiter1997long} or \ac{GRU} cell \cite{cho2014learning}. 
The initial hidden state $\mathbf{h}_0$ is usually a 0-vector. 
For a given input $\mathbf{X}^{enc} = [\mathbf{x}^{enc}_1, \mathbf{x}^{enc}_2, \ldots, \mathbf{x}^{enc}_n]$---e.g., embeddings representing words in a sentence---an encoder repeatedly applies this function, which yields an $n \times d^{enc}$ matrix $\mathbf{H}^{enc} = [\mathbf{h}^{enc}_1, \mathbf{h}^{enc}_2, \ldots, \mathbf{h}^{enc}_n]$ of $n$ hidden states of dimension $d^{enc}$.

A decoder generates output according to Equation~\ref{eq:recurrence}, where the initial hidden state is the last hidden state of the encoder $\mathbf{h}^{enc}_n$.
The predicted output at time step $t$, $\hat{o}_t$, is typically generated from the hidden state of the decoder, $h^{dec}_{t}$, by calculating a softmax over the vocabulary $V$:
\begin{equation}\label{eq:decoderOutput}
  \hat{o} = \argmax_{\mathbf{v} \in \mathbf{V}} \frac{e^{h^{dec}_{t} \cdot \mathbf{v}}}{\sum_{\mathbf{v}' \in \mathbf{V}} e^{h^{dec}_{t} \cdot \mathbf{v}'}}.
\end{equation}
Here $\mathbf{V}$ is a matrix of vector representations $v$, representing  words in the output vocabulary.
At training time, the embedding of the correct word---the word that should have been returned---is usually given as input to the recurrent cell at time step $t+1$. 

\subsection*{Attention}

An attention mechanism was proposed in \cite{bahdanau2014neural}, which gives the decoder access to the hidden states of an encoder.
Instead of using Equation~\ref{eq:recurrence} to produce a new hidden state dependent only on the input, the computation now also depends on $\mathbf{H}^{att}$, the states to attend over, typically the states of the encoder.
Following, e.g.,~\cite{vinyals2015grammar}, we have:
\begin{equation}\label{eq:attention}
  \begin{split}
    \mathbf{h}_{t}^{dec} & = g(\mathbf{x}^{dec}, \mathbf{H}^{att}, \mathbf{h}^{dec}_{t-1}) \\
    & = \mathbf{W}_{proj} \cdot \mathbf{d}_t || \hat{\mathbf{h}}_t^{dec},
  \end{split}
\end{equation}
where $||$ is the concatenation operator, $\hat{\mathbf{h}}_t^{dec} = f(\mathbf{x}^{dec}, \mathbf{h}^{dec}_{t-1} ; \theta^{dec})$ from Equation~\ref{eq:recurrence} and $\mathbf{d}_t$ is calculated from $\mathbf{H}^{att}$ by:
\begin{equation*}
  \begin{split}
    \mathbf{d}_t   & = \sum_{i=1}^n a_{t,i} ~ \mathbf{h}^{att}_i \\
    \mathbf{a}_t & = \softmax(\mathbf{u}_t) \\
    \mathbf{u}_{t,i} & = v^T \tanh(\mathbf{W}_1 \mathbf{h}^{att}_i + \mathbf{W}_2 \mathbf{h}_t^{dec}),
  \end{split}    
\end{equation*}
where $\mathbf{h}^{att}_i$ is the $i$-th state in $\mathbf{H}^{att}$ and $\mathbf{W}_1$ and $\mathbf{W}_2$ are extra parameters learned during training.
From the hidden state produced this way, output can be generated by applying Equation~\ref{eq:decoderOutput} as usual.

\begin{figure*}[t]
  \centering
  \includegraphics[width=0.9\textwidth]{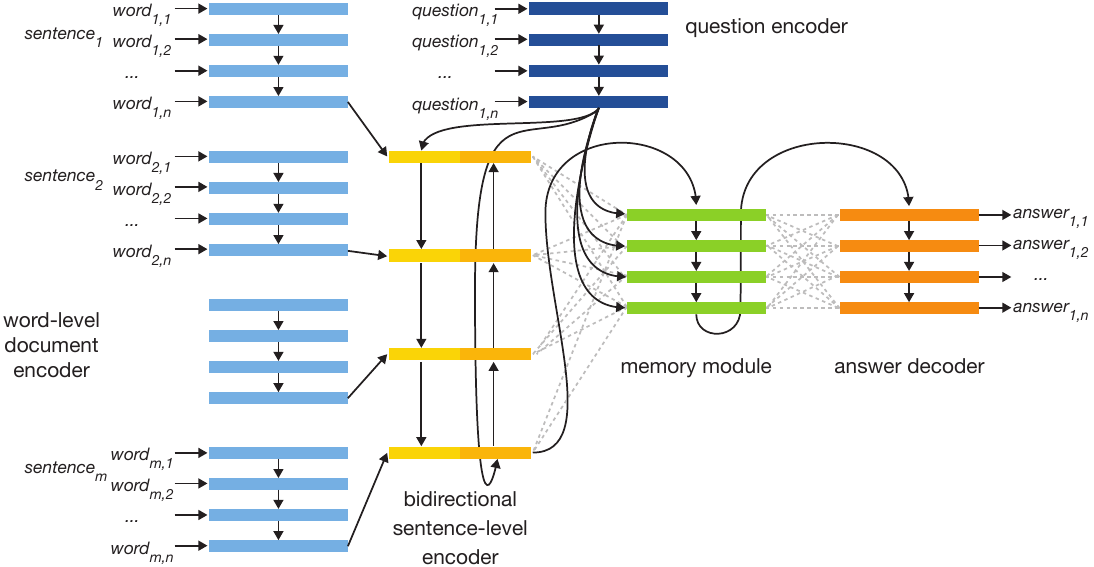}

  \bigskip
  \caption{Attentive Memory Network. Connected blocks sharing color represent RNNs. Attention is depicted by dashed lines.}
  \label{figure:amn}
  \vspace*{-\baselineskip}
\end{figure*}

\subsection{Attentive Memory Network architecture}
We now present the \acf{AMN} architecture.
\acp{AMN}, like traditional sequence-to-sequence networks, are composed of recurrent neural networks.
Their key part is a memory module, which is a recurrent network itself.
It stores memories by attending over the input document, conditioned on the question.
As can be seen from Equation~\ref{eq:attention}, the computational complexity of the attention mechanism is primarily dependent on the size of $\mathbf{H}^{att}$, the states to attend over.
To keep this matrix small, a hierarchical approach is taken, where the input is first read by a word-level document encoder, which reads word embeddings---also trained by the model---per sentence to compute sentence representations.
A sentence-level encoder iterates over these sentence embeddings to get a final document encoding.
The memory module only has access to the sentence embeddings produced by the sentence-level encoder.
For example, if the input consists of 20 sentences of 12 words each, the memory module of the \ac{AMN} attends over 20 sentence representations, rather than over 240 representations, had a non-hierarchical word-level approach been taken.

Figure~\ref{figure:amn} shows a graphical overview of the network layout.
There are two input encoders, a question encoder and a word-level document encoder.
The memory module, the green block in Figure~\ref{figure:amn}, attends over the sentence embeddings to extract relevant parts of the input, conditioned on the question.
Lastly, the answer decoder attends over the memory states, to produce the final output.
Let us turn to the details.
%

\subsubsection*{Question encoder}

For encoding the question we use a single \ac{RNN}.
For a question $Q\in \{q_1$, $q_2$, \ldots, $q_{|Q|}\}$ it produces a final state $\mathbf{h}^{que}_{|Q|}$, a vector of dimension $d^{que}$, that is used as a distributed representation of the question.

\subsubsection*{Document encoder}

To encode the document we use a hierarchical approach.
First, a word-level \ac{RNN} is used to encode sentences.
The word-level encoder is applied for every sentence individually.
The unroll length is the maximum sentence length in words.
For sentences $S\in \{s_1, s_2, \ldots, s_{|S|}\}$ the word-level encoder yields $\mathbf{H}^{wrd}$, an $|S| \times d^{wrd}$ matrix.

The sentence representations in $\mathbf{H}^{wrd}$ are read as a sequence by a sentence-level encoder.
Following, e.g., \cite{xiong2016dynamicmn}, we use a bidirectional \ac{RNN} for the sentence-level encoder, which for $|S|$ sentences and a hidden state size $d^{sen}$ yields $\mathbf{H}^{sen}$, an $|S| \times d^{sen}$ matrix.
The final state of the question encoder, $\mathbf{h}^{que}_{|Q|}$, is used as initial value of the hidden states of the sentence-level encoder.

\subsubsection*{Memory module}

The memory module consists of a single recurrent cell that produces $\mathbf{M}$, a matrix of $m$ memory representations of dimension $d^{mem}$.
The $i$-th memory $\mathbf{m}_i$ is computed conditioned on the question representation and the sentence representations, analogous to Equation~\ref{eq:attention}, as:
\begin{equation}\label{eq:memoryUpdate}
  \mathbf{m}_{i}  = g\left(\mathbf{h}_{|Q|}^{que}, \mathbf{H}^{sen}, \mathbf{m}_{i-1}\right).
\end{equation}
That is, the final representation of the question encoder $\mathbf{h}_{|Q|}^{que}$ is repeatedly provided as input to a recurrent cell, whose hidden state is computed from the memory it produced previously, $\mathbf{m}_{i-1}$, while attending over the hidden states of the sentence-level encoder $\mathbf{H}^{sen}$.

The final representation of the sentence-level document encoder $\mathbf{h}^{sen}_{|S|}$ is used to initialize the hidden state of the memory cell, $\mathbf{m}_0$.

\subsubsection*{Answer decoder}

Finally, the decoder produces an answer using Equation~\ref{eq:decoderOutput}, where $\mathbf{h}_{t}^{dec}$ is computed by attending over the memory states:
\begin{equation*}
  \mathbf{h}_{t}^{dec}  = g(\mathbf{x}^{dec}_t, \mathbf{M}, \mathbf{h}^{dec}_{t-1}).
\end{equation*}

\subsection{Efficiency}

As can be seen from Equation~\ref{eq:memoryUpdate}, the memory module is a recurrent cell itself.
In previous memory networks, the memory module passes over the input multiple times, updating memory after each pass \cite{xiong2016dynamicmn,kumar2016askma}.
The key difference in our approach is that \acp{AMN} iterate over the input only once, but \textit{attend} over it multiple times.
This is more efficient, as the attention mechanism (Equation~\ref{eq:attention}) has far less paramaters than an \ac{LSTM} or \ac{GRU} recurrent cell, which update multiple gates and an internal state at every time step.
The attention mechanism calculates a softmax over the input encodings, the number of which in our case is reduced to number of input sentences, rather than words, by the hierarchical encoder.

Additionally, the \ac{AMN} needs relatively few iterations to learn.
Details per evaluation set are provided in \S\ref{section:analysis}.


\section{Experimental setup}\label{section:experimentalSetup}

To the best of our knowledge, there is currently no conversational search data set (consisting of sequences of utterances plus questions about these utterances) on which we could evaluate \ac{AMN}. Instead we evaluate \ac{AMN} on a broad collection of more traditional machine reading datasets. Specifically, we evaluate \ac{AMN} on the 20 datasets provided by the bAbi tasks \cite{weston2015towards}, of which we use the 10k sets, version 1.2. 
The sets consist of stories, 2 to over 100 sentences in length, and questions about these stories.
The 20 sets are designed to highlight different aspects of natural language understanding like counting, deduction, induction and spatial reasoning.
As argued by \citet{kumar2016askma}, while showing the ability to solve one of the bAbi tasks is not sufficient to conclude a model would succeed at the same task on real world text data ---such as conversational search data--- it is a necessary condition.

Every dataset in the bAbi collection comes as a training set of 10,000 examples and a test set of 1,000 examples. 
We split the 10,000 training examples of each dataset into a training set---the first 9,000 examples---and a validation set---the remaining 1,000 examples---on which we tune the hyperparameters.
All text is lowercased.

We use \ac{GRU} cells \cite{cho2014learning} for all recurrent cells.  
To restrict the number of hyperparameters to tune, the same value is used for all embedding sizes, and for the state sizes of all recurrent cells.
I.e., for an embedding size $e$, we have $e = d^{que} = d^{wrd} = d^{sen} = d^{mem}$, which is either 32 or 64.
The weights of the question encoder and document word-level encoder are tied.
\ac{GRU} cells can be stacked and we experiment with 1 to 3 level deep encoder, memory, and decoder cells, the depths of which always match (i.e., if, for example, 3-level encoder cells are used, 3-level decoder cells are used).
We use a single embedding matrix for the words in the question, document and answer.
The number of memories to generate, $m$, is chosen from \{1, 2, 3\}.
Dropout is applied at every recurrent cell, the dropout probability being either 0.0 (no dropout), 0.1 or 0.2.
We optimize cross entropy loss between actual and predicted answers, using Adam \cite{kingma2014adam} as optimization algorithm and set the initial learning rate to one of \{0.1, 0.5, 1.0\}.
We measure performance every 1000 training examples.
If the loss does not improve or performance on the validation set decreases for three times in a row, the learning rate is annealed by dividing it by 2.
The maximum norm for gradients is either 1 or 5. 
The batch size is set to 50.

We implemented the \ac{AMN} in Ten\-sor\-flow \cite{tensorflow2015abadi}.
The implementation is released under an open source license and is available at \url{https://bitbucket.org/TomKenter/attentive-memory-networks-code}.

\begin{table}[t]
  \centering
  \caption{Results in terms of error rate on the bAbi 10k tasks. For comparison, results of previous work are copied from \protect\cite[MemN2N]{sukhbaatar2015endtoendmn}, \protect\cite[DNC]{graves2016hybrid}, \protect\cite[DMN+]{xiong2016dynamicmn}, and \protect\cite[EntNet]{henaff2016tracking}.}
  \begin{tabular}{l@{} c @{~~~} c @{~~~} c @{~~~} c @{~~~} c}
    \toprule
  Dataset & MemN2N & DNC & DMN+ & EntNet & AMN \\
  \midrule
  single supporting fact & 0.0  & 0.0  & 0.0  & 0.0 & 0.0 \\ 
  two supporting facts   & 0.3  & 0.4  & 0.3  & 0.1 & 4.1 \\ 
  three supporting facts & 2.1  & 1.8  & 1.1  & 4.1 & \smash{\llap{2}}9.1 \\ 
  two arg relations      & 0.0  & 0.0  & 0.0  & 0.0 & 0.0 \\ 
  three arg relations    & 0.8  & 0.8  & 0.5  & 0.3 & 0.7 \\ 
  yes-no questions       & 0.1  & 0.0  & 0.0  & 0.2 & 0.2 \\ 
  counting               & 2.0  & 0.6  & 2.4  & 0.0 & 3.1 \\ 
  lists sets             & 0.9  & 0.3  & 0.0  & 0.5 & 0.3 \\ 
  simple negation        & 0.3  & 0.2  & 0.0  & 0.1 & 0.0 \\ 
  indefinite knowledge   & 0.0  & 0.2  & 0.0  & 0.6 & 0.1 \\ 
  basic coreference      & 0.1  & 0.0  & 0.0  & 0.3 & 0.0 \\ 
  conjunction            & 0.0  & 0.0  & 0.0  & 0.0 & 0.0 \\ 
  compound coreference   & 0.0  & 0.1  & 0.0  & 1.3 & 0.0 \\ 
  time reasoning         & 0.1  & 0.4  & 0.2  & 0.0 & 3.6 \\ 
  basic deduction        & 0.0  & 0.0  & 0.0  & 0.0 & 0.0 \\ 
  basic induction        & \smash{\llap{5}}1.8 & \smash{\llap{5}}5.1 & \smash{\llap{4}}5.3 & 0.2 & \smash{\llap{4}}5.4 \\ 
  positional reasoning   & \smash{\llap{1}}8.6 & \smash{\llap{1}}2.0 & 4.2  & 0.5 & 1.6 \\ 
  size reasoning         & 5.3  & 0.8  & 2.1  & 0.3 & 0.9 \\ 
  path finding           & 2.3  & 3.9  & 0.0  & 2.3 & 0.3 \\ 
  agents motivations     & 0.0  & 0.0  & 0.0  & 0.0 & 0.0 \\ 
  \midrule
  number of tasks solved & 18 & 18 & 19 & 20 & 18 \\
  \bottomrule
  \end{tabular}
  \label{table:mainResults}
\end{table}


\begin{table}[t]
  \centering
  \caption{Hyperparameter values for the minimal \acp{AMN} that were fastest in achieving best performance on the validation set. The size refers to both size of embeddings and hidden states. The last column lists the number of batches needed.}
  \begin{tabular}{l c @{~} c @{~} c @{~} c @{~} c@{}}
  \toprule
  Dataset & size & \#\thinspace layers & \#\thinspace mem  & \multicolumn{1}{c}{\#\thinspace batches} \\
  \midrule
  single supporting fact & 32  & 1 & 1 & 1,000  \\ 
  two supporting facts   & 64  & 2 & 3 & \smash{\llap{1}}2,200 \\ 
  three supporting facts & 64  & 2 & 3 & \smash{\llap{1}}4,000 \\ 
  two arg relations      & 32  & 1 & 1 & 1,200  \\ 
  three arg relations    & 32  & 1 & 2 & 3,000  \\ 
  yes-no questions       & 32  & 1 & 1 & 3,800  \\ 
  counting               & 32  & 1 & 3 & 5,000  \\ 
  lists sets             & 32  & 1 & 1 & 4,400  \\ 
  simple negation        & 32  & 1 & 2 & 3,200  \\ 
  indefinite knowledge   & 32  & 1 & 1 & 3,800  \\ 
  basic coreference      & 32  & 1 & 2 & 1,400  \\ 
  conjunction            & 32  & 1 & 1 & 1,200  \\ 
  comp coreference       & 32  & 1 & 1 & \smash{\llap{1}}0,000 \\ 
  time reasoning         & 64  & 2 & 1 & 6,000  \\ 
  basic deduction        & 32  & 1 & 1 & 2,200  \\ 
  basic induction        & 64  & 1 & 2 & \smash{\llap{1}}0,200 \\ 
  positional reasoning   & 32  & 1 & 3 & 6,200 \\ 
  size reasoning         & 32  & 1 & 3 & 2,400 \\ 
  path finding           &64   & 1 & 1 & \smash{\llap{1}}3,000 \\ 
  agents motivations     & 32  & 1 & 3 & 3,600 \\ 
  \bottomrule
  \end{tabular}
  \label{table:speedTest}
  \vspace*{-\baselineskip}
\end{table}

\section{Results and analysis}


\begin{figure*}[t]
  \centering
\begin{subfigure}[t]{.3\textwidth}
  \centering
  \includegraphics[width=\textwidth,clip,trim=0mm -2.9mm 0mm 0mm,valign=t]{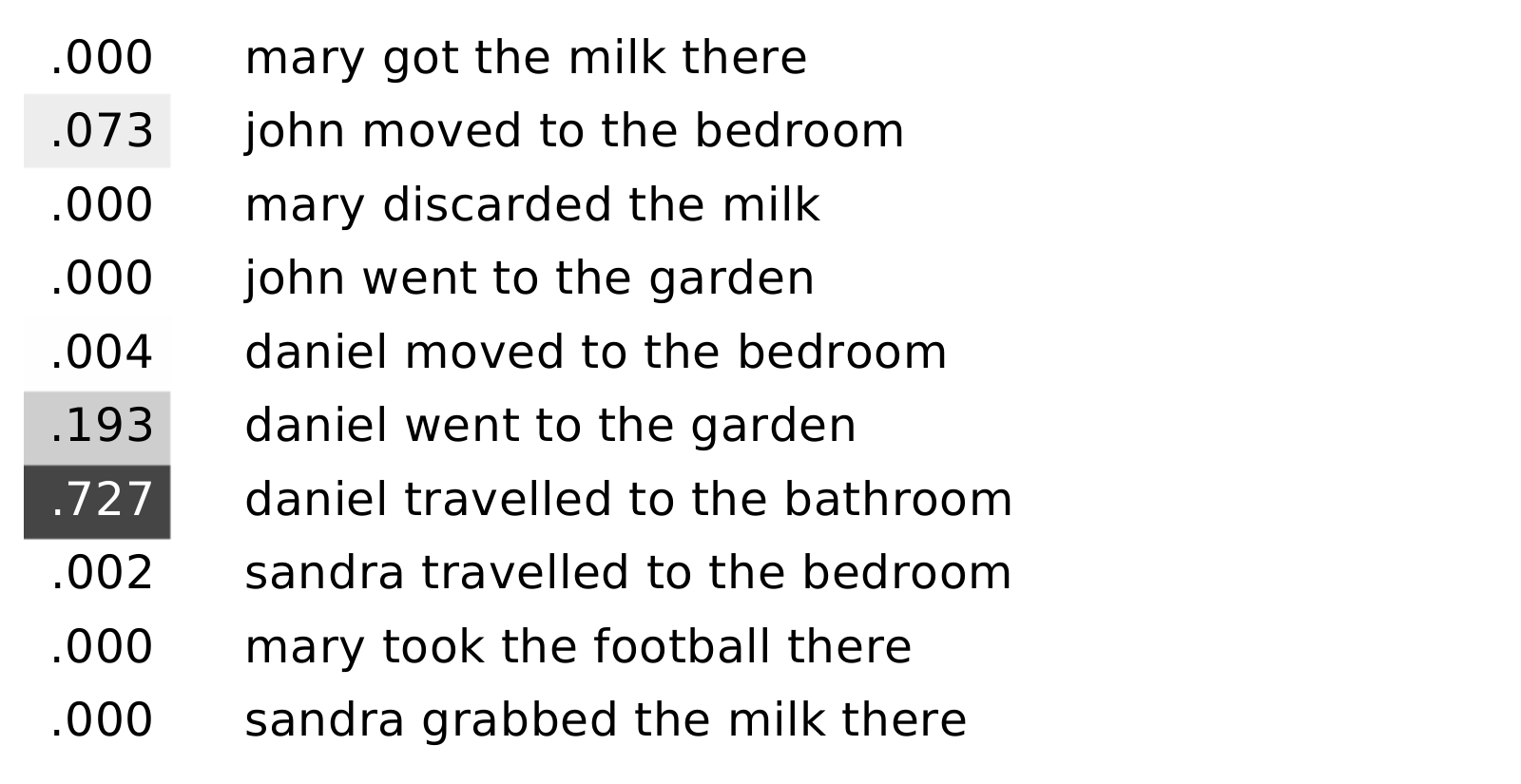}
  \caption{Dataset: yes-no questions, question: `is daniel in the bedroom?', prediction: `no', ground truth: `no'.}
  \label{figure:qa6}
\end{subfigure}
\quad
\begin{subfigure}[t]{.31\textwidth}    
  \centering
  \includegraphics[width=\textwidth,valign=t]{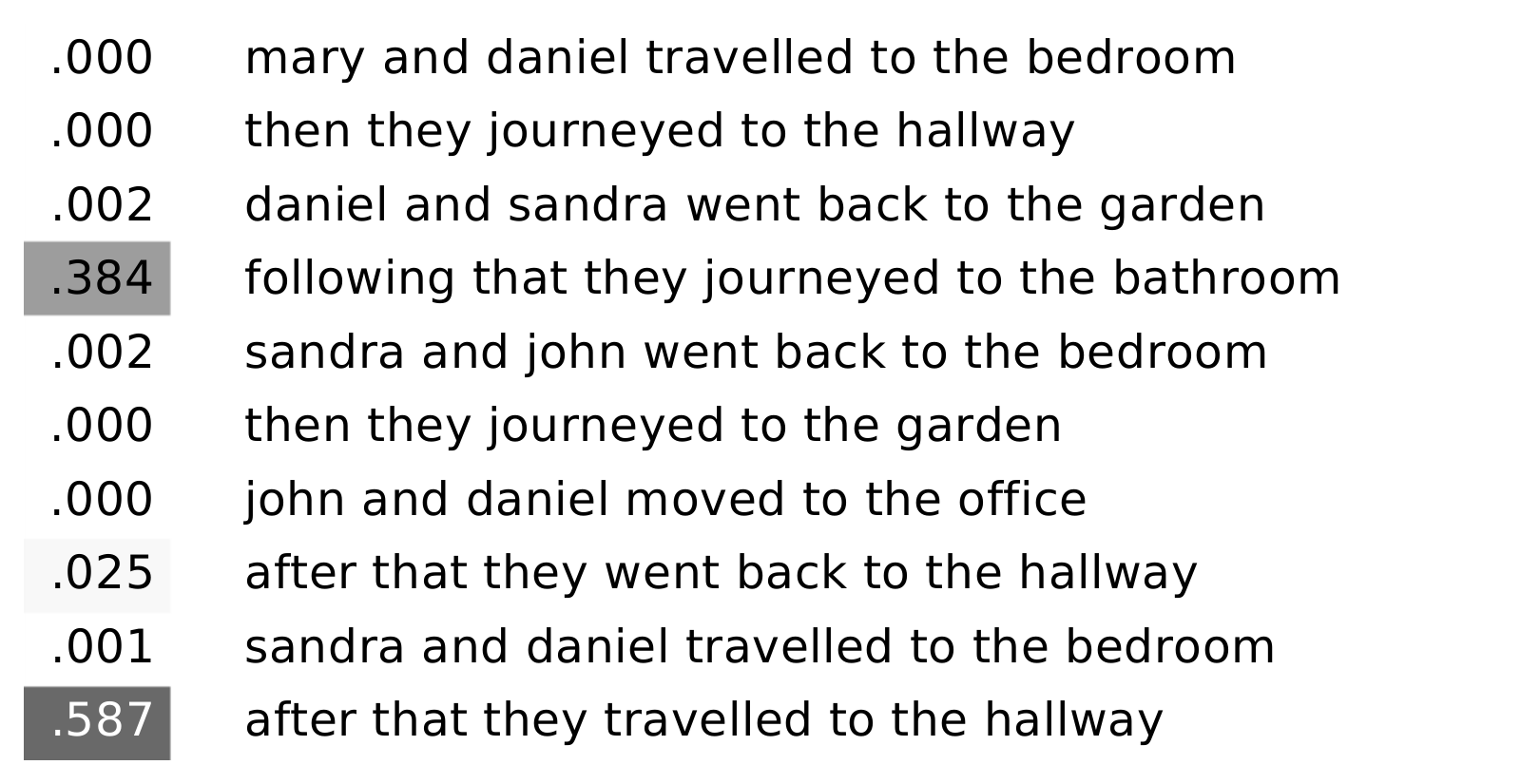}
  \caption{Dataset: compound coreference, question: `where is daniel?', prediction: `hallway', ground truth: `hallway'.}
  \label{figure:qa13}
\end{subfigure}
\quad
\begin{subfigure}[t]{.31\textwidth}
  \centering
  \includegraphics[width=\textwidth,clip,trim=0mm -4mm 0mm 0mm,valign=t]{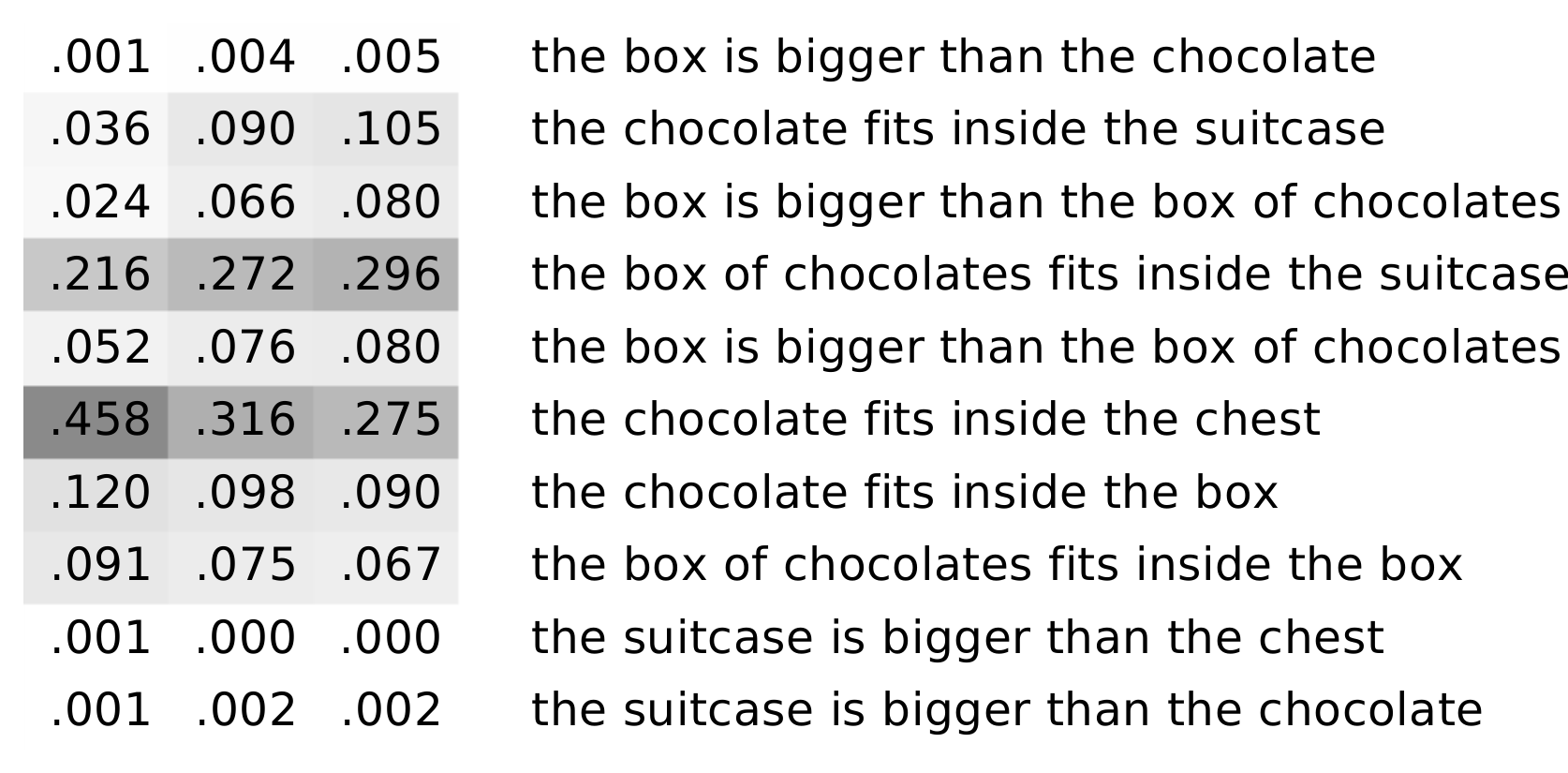}
  \caption{Dataset: size reasoning, question: `is the suitcase bigger than the chocolate?', prediction: `yes', ground truth: `yes'.}
  \label{figure:qa18} 
\end{subfigure}
\\
\begin{subfigure}[t]{.31\textwidth}
  \centering
  \includegraphics[width=\textwidth,trim=0mm -23.7mm 0mm 0mm,valign=t]{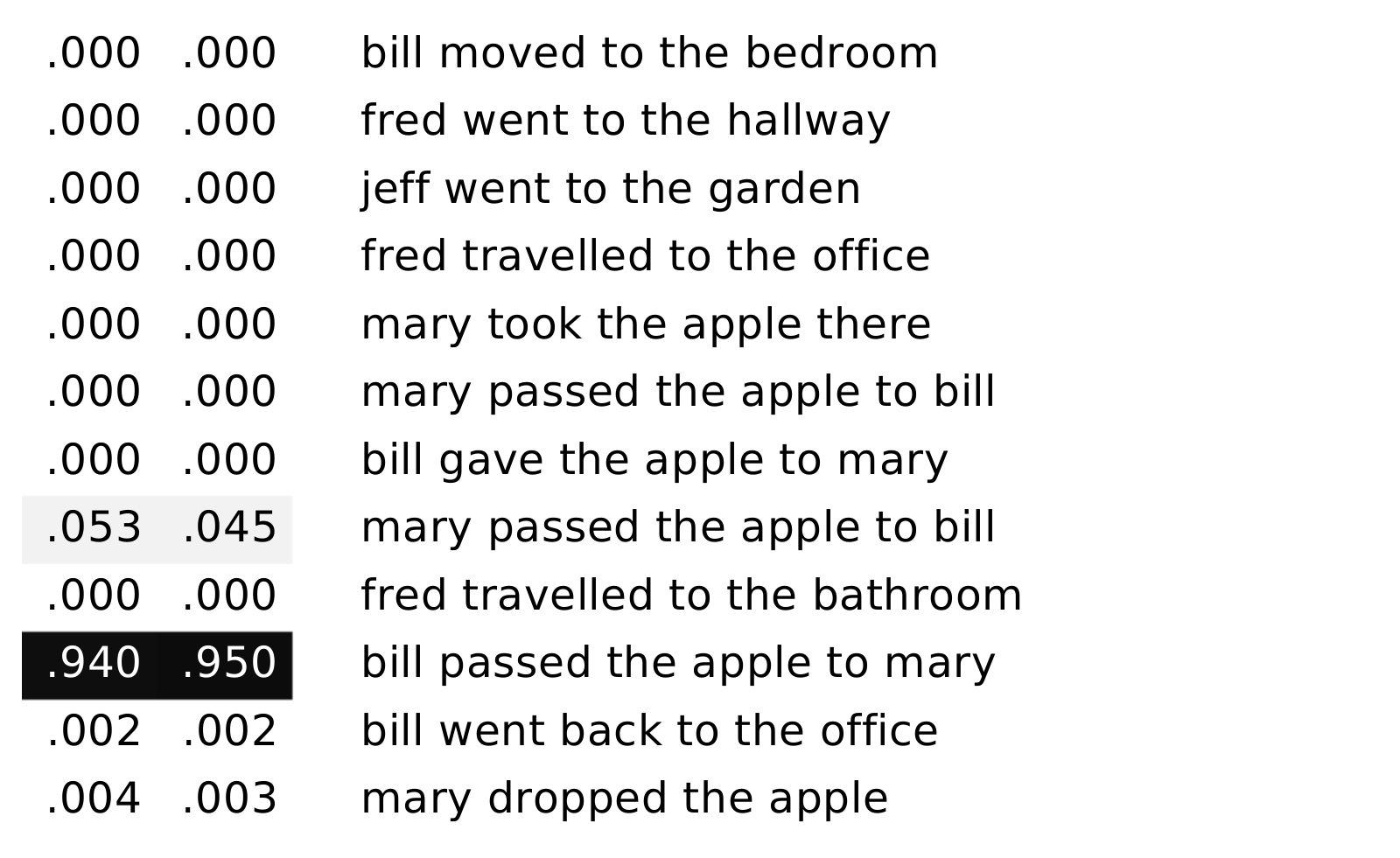}
  \caption{Dataset: three arg relations, question: `what did bill give to mary?', prediction: `apple', ground truth: `apple'.}
  \label{figure:qa5}
\end{subfigure}
\quad
\begin{subfigure}[t]{.31\textwidth}    
  \centering
  \includegraphics[width=\textwidth,clip,trim=0mm 0mm 0mm 0mm,valign=t]{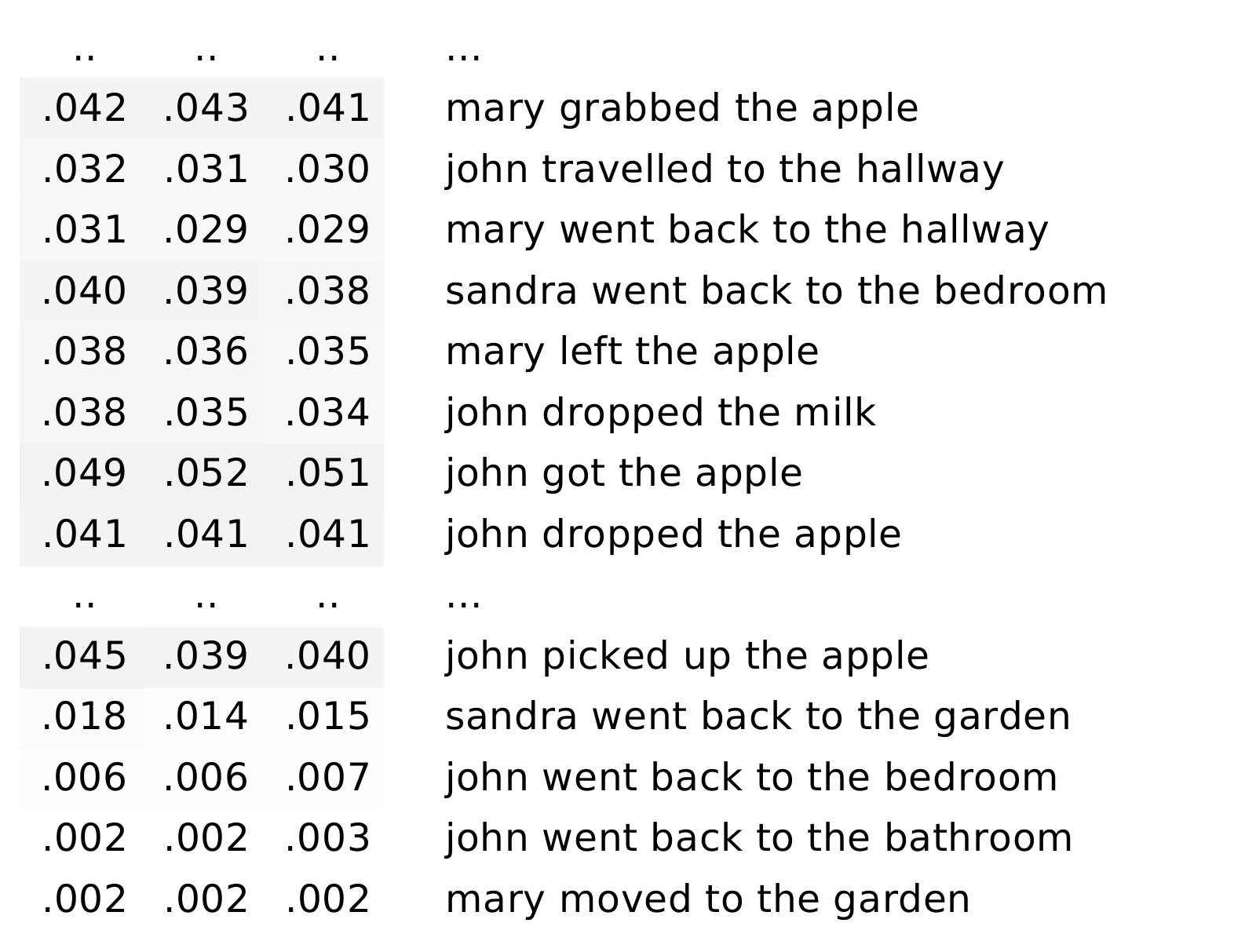}
  \caption{Dataset: three supporting facts, question: `where was the apple before the bathroom?', prediction: `garden', ground truth: `bedroom'.}
  \label{figure:qa3_1}
\end{subfigure}
\quad
\begin{subfigure}[t]{.31\textwidth}    
  \centering
  \includegraphics[width=\textwidth,clip,trim=0mm 0mm 0mm 0mm,valign=t]{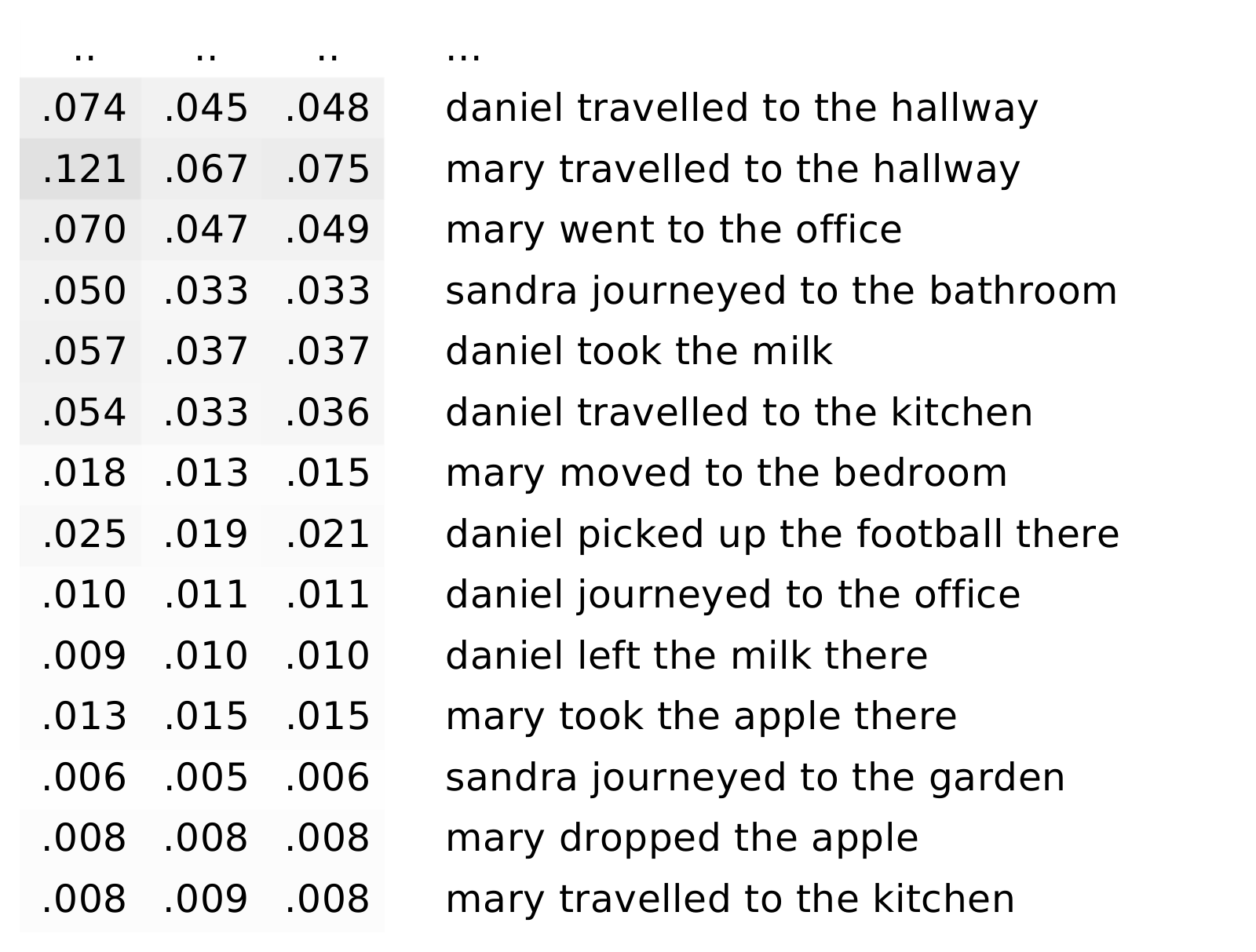}
  \caption{Dataset: three supporting facts, question: `where was the milk before the office?', prediction: `hallway', ground truth: `kitchen'.}  
  \label{figure:qa3_2}
\end{subfigure}

  \vspace*{-.5\baselineskip}
  \caption{Attention visualizations. The attention is visualized per memory step. Every column represents a memory step, and adds up to 1 (allowing for rounding errors), except in the last two examples where some (irrelevant) sentences were left out. Although some stories in the dataset are over 100 sentences in length, short examples were picked here, for brevity.}
  \label{figure:attentionVisualizations}
\end{figure*}

We present the results of the experiments described in \S\ref{section:experimentalSetup} and provide an analysis of the results. 

\subsection{Main results}

Table~\ref{table:mainResults} lists the results of our \acf{AMN} on the 20 bAbi 10k datasets, together with results of previous approaches.
Following~\cite{weston2015towards}, we consider a dataset solved if the error rate is less than 5\%.

As can be seen from the Table~\ref{table:mainResults}, \ac{AMN} solves 18 of the 20 datasets.
This is particularly noteworthy given the fact that it is a general framework, not catered towards tracking entities (as in \cite{henaff2016tracking}).
Moreover, the \ac{AMN} needs an order of magnitude fewer computation steps than previous memory network architectures used for these tasks~\cite{kumar2016askma,xiong2016dynamicmn} as it only reads the input once.

There are two tasks the \ac{AMN} does not solve.
The \textit{basic induction} set proves to be hard for the \ac{AMN}, as it does for most other networks.
More interestingly, the \textit{three supporting facts} sets is problematic as well.
This dataset has the longest documents, sometimes over 100 sentences long.
Analysis of the results, see below for examples, shows that the probability mass of the attention vectors of the memory module is much more spread out across sentences then it is in other sets.
That is, the network struggles to keep its attention focused.

The results in Table~\ref{table:mainResults} show that the \ac{AMN} can solve a wide variety of machine reading tasks and that it behaves different from other memory networks.

\subsection{Analysis}\label{section:analysis}

We analyze the hyperparameter settings used to produce the results in Table~\ref{table:mainResults} and provide examples of the inner workings of the attention mechanism of the memory module.

\subsubsection*{Hyperparameters and speed of convergence}

Table~\ref{table:speedTest} lists the hyperparameter values for the smallest \acp{AMN} that achieve the best performance on the validation set, with fewest training examples.
Here, \textit{smallest network} refers to the size of the network in terms of embedding size and number of memories.
The last column lists the number of batches needed.
As can be seen from Table~\ref{table:speedTest}, \acp{AMN} can learn fast.
As an example, it needs only 5 epochs to solve the first dataset: there are 10k examples---1,000 batches of 50 examples = 50k examples = 5 epochs.
This is in contrast to the 100 epochs reported in \cite{sukhbaatar2015endtoendmn} and 256 epochs listed as a maximum in \cite{kumar2016askma}.

Interestingly, adding depth to a network by stacking \ac{GRU} cells was helpful in only 3 out of 20 cases.

\subsubsection*{Result analysis}

Figure~\ref{figure:attentionVisualizations} shows a visualization of the attention vectors of the memory module.
The attention is visualized per memory step.
Although some stories in the dataset are over 100 sentences in length, short examples were picked here, for reasons of brevity.
Every column represents a memory step, and the values per memory step add up to 1 (barring rounding errors).

Figure~\ref{figure:qa6} shows an example where one memory step is used.
The attention focuses on the last time Daniel, the person the question is about, is mentioned.
Interestingly, the second sentence also gets some attention, presumably because the bedroom, which features in the question, is being referred to.
A particularly striking detail is that---correctly---nearly no attention is paid to the fifth sentence, although it is almost identical to the question.

In Figure~\ref{figure:qa13}, attention is highest for sentences in which the person being asked about is referred to.
This is especially noteworthy, as the reference is only by a personal pronoun, which moreover refers to two people. 

For the \textit{size reasoning} dataset, three memory steps were needed (see Table~\ref{table:speedTest}).
An example is shown in Figure~\ref{figure:qa18}.
The first memory step mistakenly focuses on the sixth sentence about the chest.
Gradually, however, the memory module recovers from this error, and attention shifts to the fourth sentence about the suitcase.

Figure~\ref{figure:qa5} shows the ability of the network to focus only on relevant parts.
Although the seventh and tenth sentence are nearly identical, it is the last sentence that matters, and it is this sentence the network attends to almost solely.
Curiously, the two memory steps attend to the same sentences, which is consistently the case for this dataset.
This might indicate that a single memory step could suffice too.
Indeed, experiments show that on some datasets networks with fewer memory steps achieve the same or nearly the same performance as bigger networks, but take longer to reach it.
The extra memory steps might serve as extra training material.

The last two cases, Figure~\ref{figure:qa3_1} and \ref{figure:qa3_2}, are from the \textit{three supporting facts} dataset that the model could not solve.
What stands out immediately is the fact that the attention is much more spread out than in other cases.
This is the case throughout the entire dataset.
It shows that the model is confused and fails to learn what is relevant.
In Figure~\ref{figure:qa3_1} just reading the last five sentences would have been enough.
The model does seem to capture that John picked up the apple, but only very weakly so.
The crucial sentence, third form the end, is the sentence the model pays least attention to.
Figure~~\ref{figure:qa3_1} shows the model being even more confused.
It starts out by attending mostly to Mary, who has nothing to do with the story.
The sentences that do matter, again, get very little attention.

Overall, these examples indicate that, when the \ac{AMN} learns to solve a task, its memory module is very decisive in paying attention to the relevant parts of the input and ignoring the rest.  


\section{Conclusion}

As search becomes more conversational, the machine reading task, where a system is able to answer questions against prior utterances in a conversation, becomes a highly relevant task for \ac{IR}.
We introduced \acfp{AMN}, efficient end-to-end trainable memory networks with a hierarchical input encoder.
\acp{AMN} perform nearly as well as existing machine reading algorithms, with less computation.
Analysis shows they typically need only a few epochs to achieve optimal performance, making them ideally suited for \ac{IR}'s high efficiency settings.
Our findings indicate that a straightforward architecture like the \ac{AMN} is sufficient for solving a wide variety of machine reading tasks.

The bAbi datasets provide an ideal test bed for machine reading algorithms as the tasks and evaluation are well-defined.
However, it would also be interesting to test the performance of \acp{AMN} on bigger datasets, with more varied and noisier problems, especially ones that are directly derived from conversational search scenarios.

Memory networks have also been applied in settings where external knowledge is available, in particular in the form of key-value pairs \cite{miller2016keyvaluemn}.
Although this setting is different from the machine reading setting, it would be interesting to see how \acp{AMN} could be applied here.
Finally, in a conversational setting involving multiple actors, it would be challenging for the memory module to attend to the utterances of the right actor at the right time.
A richer attention-like mechanism seems to be needed.
One that allows a decoder to attend to specific parts of the input, including the utterances produced by the system itself, conditioned on whose utterances are being referred to.

\section*{Acknowledgments}
We would like to thank Nikos Voskarides of the University of Amsterdam for valuable feedback on an earlier version of the manuscript, and Llion Jones and Daniel Hewlett of Google Research for many inspiring discussions on topics related to the work in this paper.

This research was supported by
Ahold Delhaize,
Amsterdam Data Science,
the Bloomberg Research Grant program,
the Criteo Faculty Research Award program,
the Dutch national program COMMIT,
Elsevier,
the European Community's Seventh Framework Programme (FP7/2007-2013) under
grant agreement nr 312827 (VOX-Pol),
the Microsoft Research Ph.D.\ program,
the Netherlands Institute for Sound and Vision,
the Netherlands Organisation for Scientific Research (NWO)
under pro\-ject nrs
612.001.116, 
HOR-11-10, 
CI-14-25, 
652.\-002.\-001, 
612.\-001.\-551, 
652.\-001.\-003, 
and
Yandex.
All content represents the opinion of the authors, which is not necessarily shared or endorsed by their respective employers and/or sponsors.

\bibliographystyle{ACM-Reference-Format}
\bibliography{attentive_memory_networks} 

\end{document}